\pdfoutput=1
\documentclass[11pt]{article}
\usepackage{kotex}
\usepackage{acl}

\usepackage{times}
\usepackage{latexsym}
\usepackage{graphicx, wrapfig}
\usepackage{ulem}
\newcommand{\changeline}[1]{\textcolor{black}{#1}}
\newcommand{\changelines}[1]{\textcolor{black}{#1}}
\newcommand{\changeliness}[1]{\textcolor{black}{#1}}
\newcommand{\finalEdit}[1]{\textcolor{black}{#1}}

\newcommand{\aclStyle}[1]{\textcolor{black}{#1}}

\usepackage[T1]{fontenc}
\usepackage[utf8]{inputenc}

\usepackage{microtype}

\title{Towards the Human Global Context: \\Does the Vision-Language Model Really Judge Like a Human Being?}

\author{Sangmyeong Woh\textsuperscript{$\dagger$}, Jaemin Lee\textsuperscript{$\dagger$}, Ho joong Kim \and Jinsuk Lee\textsuperscript{*} \\
Testworks Inc., Seoul, Republic of Korea \\
\texttt{\{wohsang, jaemin, hjk1221, jslee\}@testworks.co.kr} \\}
\begin{document}
\maketitle
\def\thefootnote{$\dagger$ }\footnotetext{These authors contributed equally.}
\def\thefootnote{* }\footnotetext{Corresponding author.}

\begin{abstract}
\changelines{As computer vision and NLP make progress, Vision-Language (VL) is becoming an important area of research. Despite the importance, evaluation metrics of the research domain is still at a preliminary stage of development. In this paper, we propose a quantitative metric \textit{"Equivariance Score"} and evaluation dataset \textit{"Human Puzzle"} to assess whether a VL model is understanding an image like a human.  We observed that the VL model does not interpret the overall context of an input image but instead shows biases toward a specific object or shape that forms the local context. We aim to quantitatively measure a model’s performance in understanding context. To verify the current existing VL model’s capability, we sliced the original input image into pieces and randomly placed them, distorting the global context of the image. Our paper discusses each VL model’s level of interpretation on global context and addresses how the structural characteristics influenced the results.}\\
\end{abstract}

\section{Introduction}
\begin{figure*}[t]
    \centering
    \includegraphics[width=\linewidth]{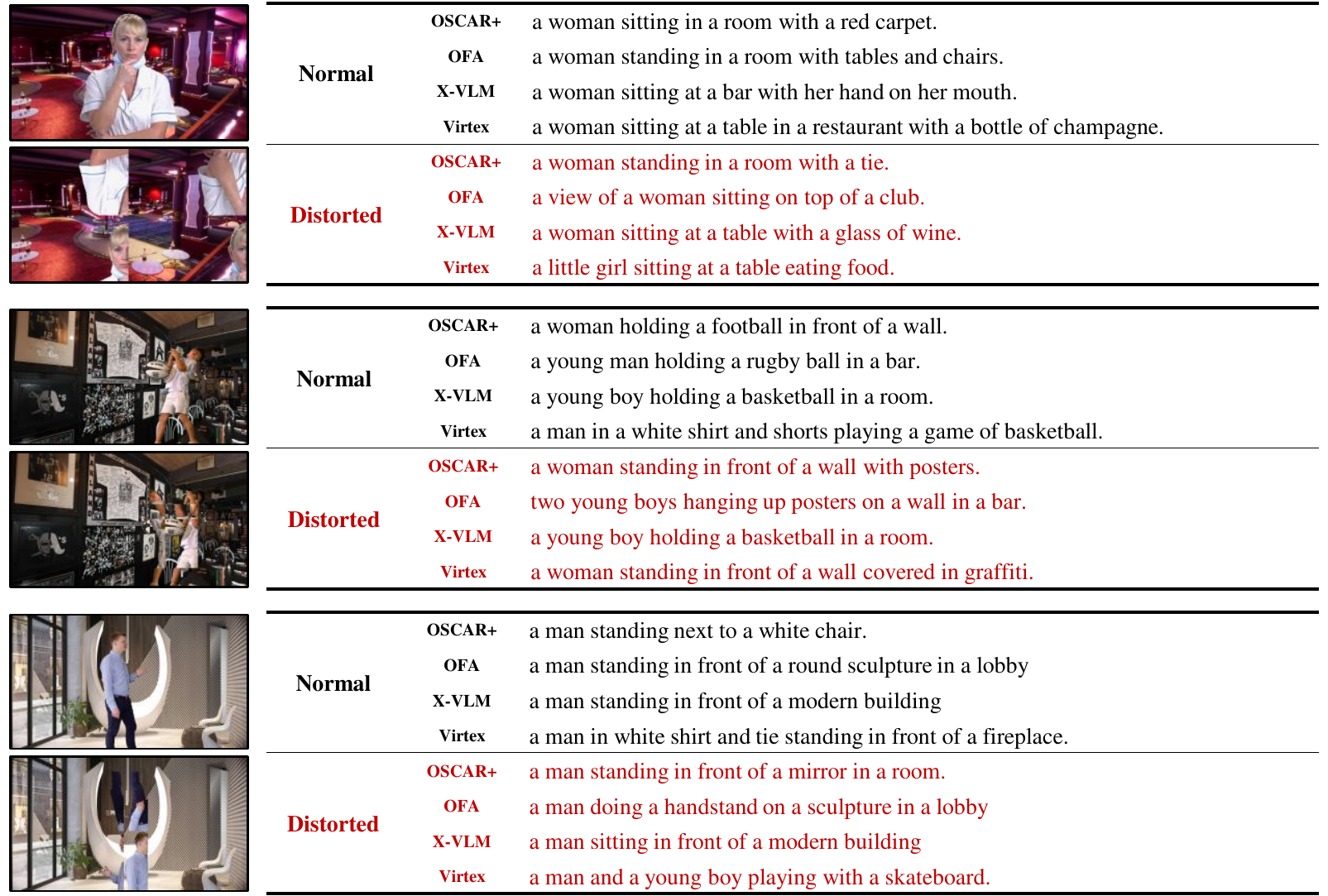}
    \caption{\textbf{Captioning result of each model in \textit{“Human Puzzle”} dataset:} Can you recognize which model is better at handling global context? In general, OSCAR+ performs better in most VL tasks, but this does not guarantee that it will excel in treating context information as well.}
    \label{fig:fig1}
\end{figure*}

Vision-Language task interprets the context of the given image and attempts to represent it \changelines{in a human language}. Image captioning~\cite{wang2022unifying, zeng2021multi, wang2021simvlm, pan2020x} for example, analyzes context (relations between objects, situational background, etc.) to construct a sentence that adequately explains the given image. In a similar domain, VQA~\cite{antol2015vqa} uses multi-modal to generate adequate answers to the text (question) associated with the image. As image-based text generation models infer results from fusing computer vision domain and NLP domain, majority of VL models are classified into “vision task”~\cite{he2016deep, ren2015faster}, which interprets the given image, and “text task” which generates the associated texts. Most common \changelines{methodologies} used is the 2-stage approach, where the vision model first extracts the features from the image that are easier for the language model to interpret and train from. In the case of OSCAR~\cite{li2020oscar}, the maximum number of objects are detected from the object detection model and are passed on to a language model with the object's class name. In the case of X-VLM~\cite{zeng2021multi}, inter-object relation was used in training for better information sharing between image-text models. In the case of OFA~\cite{wang2022unifying}, conveying information that is much more comprehensible for language models was the focus to enhancing performance. As stated above, there has been a growing body of research that explores how recent VL models can better interpret images and deliver critical information to the language model. 

Then what evaluation method would be appropriate to \changeline{compare which structure is better at Vision-Language process?}  Perplexity~\cite{jelinek1977perplexity} at language generation? BLEU~\cite{papineni2002bleu} \changeline{score?} CIDEr~\cite{vedantam2015cider}? These are all valid metrics to measure VL model’s performance, but they all do not properly evaluate all aspects of VL model. \changeline{Figure~\ref{fig:fig1} depicts results of captioning from several VL models. Each model came up with its own answer for the image of a distorted human body. We human beings can easily recognize \changelines{differences} between normal and distorted images, so VL models are expected to be equivariant (exlpained in section~\ref{sec:2.4}) as well but that is not the case. In view of the captioning results, all VL models seem to concentrate more on local-context without considering distorted human body(a.k.a global context of human body). \changelines{Despite VL models being invariant(non-equivariant) with captioning, each model has different understandings in global context of human body. This is obvious because they all have their own feature extraction stage on image. } To perform evaluation on such VL models, we propose \textit{Equivariance Score} and \changelines{dataset named as \textit{\textbf{“Human Puzzle”}}}. As a result, our contributions are the following:}\\

1. We illustrate how Vision-Language models are biased towards local context instead of global context.\\

2. We present an evaluation dataset and metrics to measure global context interpretation.
\\

\begin{figure*}[t]
  \centering
  \includegraphics[width=\linewidth]{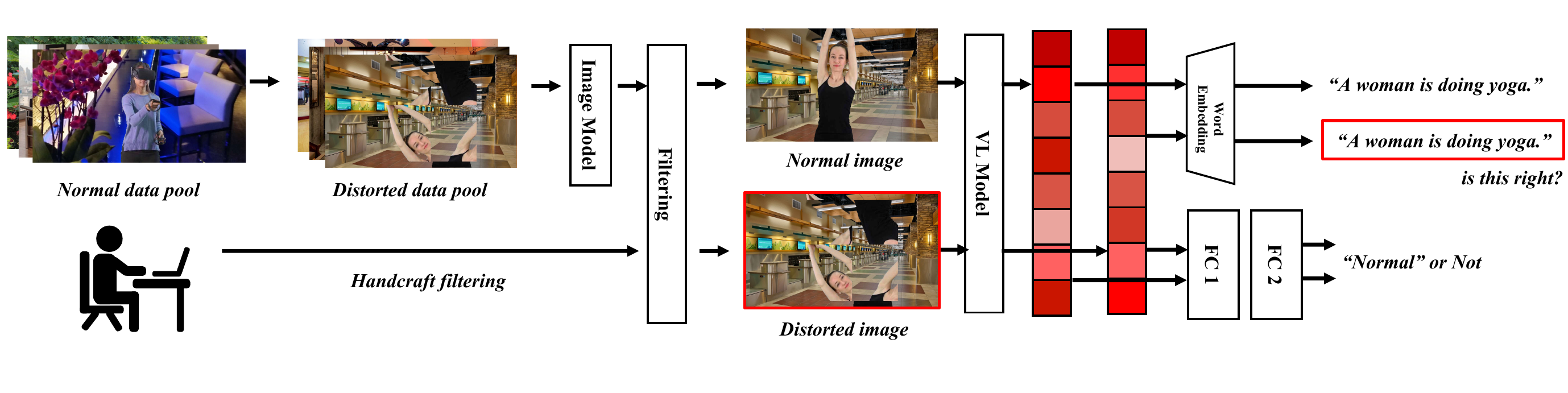}
  \caption{A pipeline from construction of \textit{Human Puzzle dataset} to  train the dataset.}
  \label{fig:fig2}
\end{figure*}

\section{Related Works}
\subsection{Vision-Language Networks}
\changelines{As vision and language studies make progress, importance in the Vision-Language model is also steadily rising. In recent years, OSCAR+, which replaces image interpreting stage of OSCAR with VinVL~\cite{zhang2021vinvl}, has demonstrated notable performance.} Through the method of \changelines{placing} an object's label \changelines{with} its extracted feature from the image \changelines{ and} sending it to the transformer-encoder structure, OSCAR+ has achieved state-of-the-art performances in numerous VL tasks and quickly became a dominant model in the \changelines{VL} field. In \changelines{a} similar fashion, X-VLM also extracts features from images and trains them to language model, but through many layers of contrastive learning on inter-object relations, the model sought to understand not only the single object context but also the relationship between objects in general. \changelines{OFA also tried to extract information from the image and made the language model interpret the results.} In addition, unique tokens were allocated to help the language model have a better perception of the extracted information, and this kind of process played a pivotal role in narrowing the gap between the language and image models. In Vision Language Processing, researchers are still studying various models to find a better way of extracting information from images.\\

\subsection{Image Networks}
\textit{Human Puzzle} dataset was first constructed by producing large amounts of data with distorted context of human body. We then filtered by choosing ones where the vision model could not identify correctly from the undistorted. Vision models utilized in this process are ResNet~\cite{he2016deep}, Vision Transformer~\cite{dosovitskiy2020image}, and Swin Transformer~\cite{liu2021swin}, which were selected based on prevalence. 
In ResNet, skip-connection was applied to increase the model's depth without vanishing gradient problem. As a result, stable \changelines{training} while maintaining deep networks became possible, propelling ResNet to be a dominant backbone in fields of vision. Vision Transformer is a model that integrated the structure of a transformer, a model that made compelling breakthroughs in the field of NLP. By partitioning images into patches and treating them like text tokens, Vision Transformer shows state-of-the-art performance comparable to conventional CNN-based models regarding classification. Swin Transformer is a model derived from CNN’s approach to overcome shortcomings of Transformer and likewise exhibits state-of-the-art performance in classification.\\

% \subsection{Global-Context in Vision-Language}
% There have been many preceding attempts~\cite{huang2019ccnet, seo2021sequential, lee2022exploiting} to analyze global context in the field of vision. In prior studies, DeepLabV3~\cite{chen2017rethinking} and CCNet~\cite{huang2019ccnet} applied ASPP (Atrous Spatial Pyramid Pooling) and CC-attention (Criss-Cross attention) to widen feature’s receptive field to utilize global context, reporting high performances as a result. 

% There also have been attempts to interpret global context in VL models through explicit or implicit methods, generally directed towards extracting additional information from the image. The use of inter-object relation information for training in X-VLM and translation of extracted detection information for better comprehensibility to language models can all be seen as part of an attempt to make models have a better grasp of understanding global context. In the case of OSCAR, the most dominant model in the field of VL however, it has been observed that intentionally corrupting an object's positional information had no impact on its performance. Based on these findings, we suggest that OSCAR relatively lacks in its ability to interpret global context.\\

\subsection{Interpretation of Global Context}

There have been many preceding attempts~\cite{huang2019ccnet} to analyze global context in the field of vision. In prior studies, DeepLabV3~\cite{chen2017rethinking} and CCNet~\cite{huang2019ccnet} applied ASPP (Atrous Spatial Pyramid Pooling) and CC-attention (Criss-Cross attention) to widen feature’s receptive field to utilize global context, reporting high performances. 

In Vision Language Processing, there also has been implicit attempts to interpret global context in images. VL models are generally directed towards extracting additional information from the image such as specifying object relation with natural languages, giving positional information of detected object to Language Model, translating image features into natural language, and so on. These can all be seen as an attempt to make models have a better grasp of understanding global context. In the case of OSCAR+, the most dominant model in the field of VL \changelines{that tried to convey positional information with features and labels of detected object} however, has been observed that intentionally corrupting an object's positional information has no impact on its performance. Based on these findings, we guess that OSCAR+ has less consideration in interpretation of global context . \\

\subsection{Equivariance and Invariance}
\label{sec:2.4}

\begin{figure}[h]
  \centering
  \includegraphics[width=\linewidth]{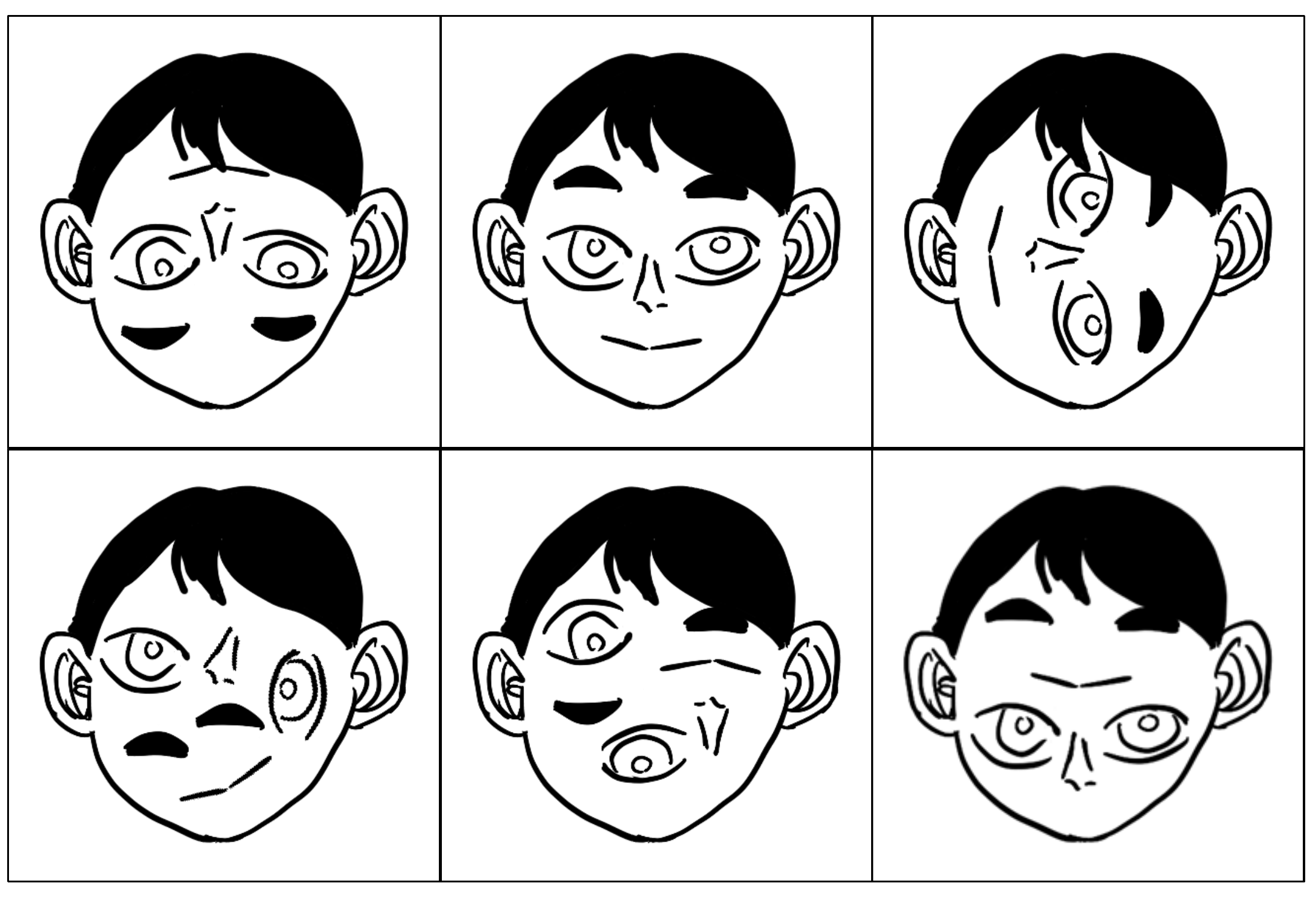}
  \caption{Face images with same local context, but different global context. In ideal model equivariance, all models should give different results for each image.}
  \label{fig:2.4}
\end{figure}

\begin{figure*}[t]
  \centering
  \includegraphics[width=\linewidth]{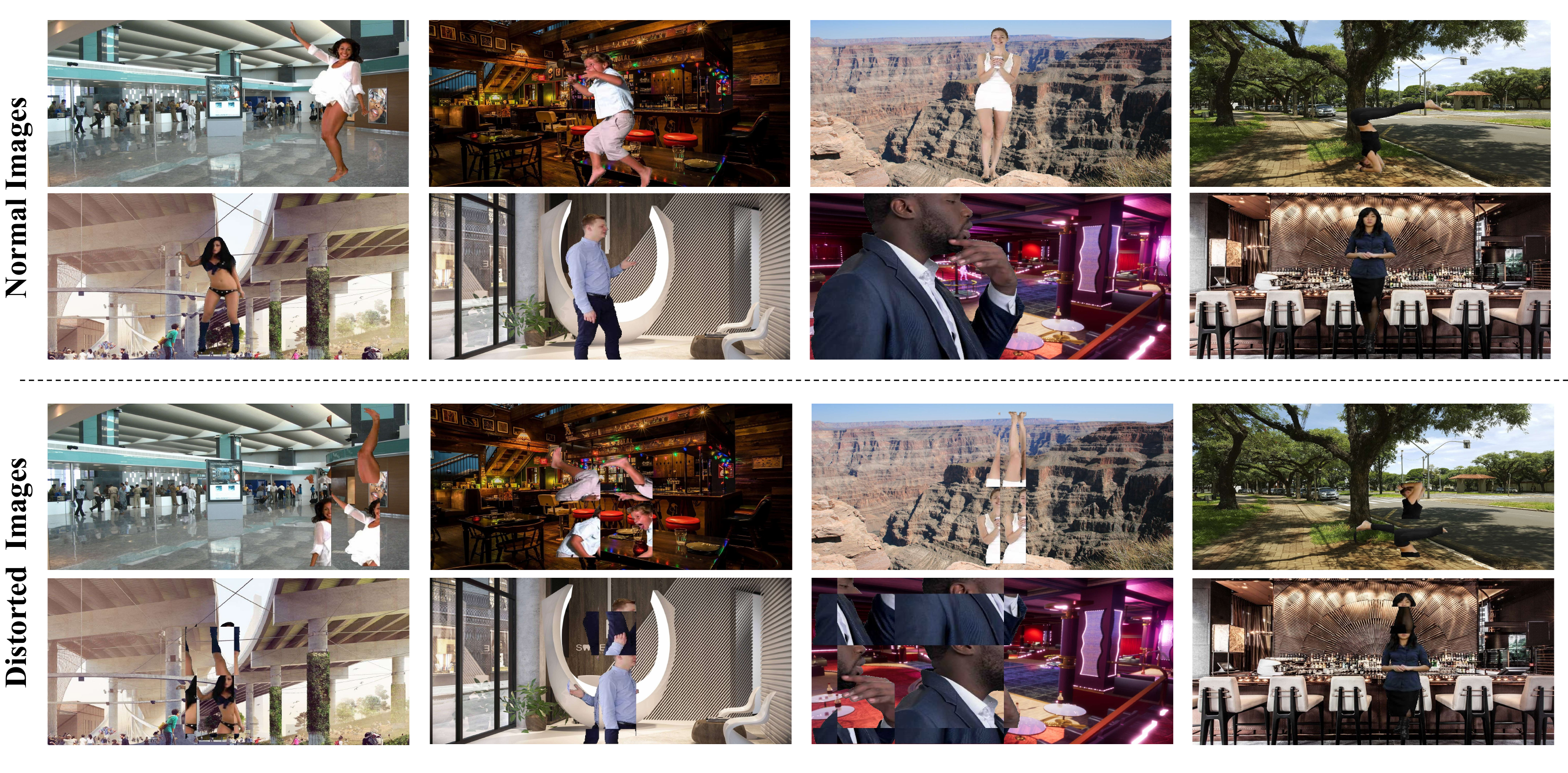}
  \caption{\textbf{Samples of \textit{Human Puzzle dataset}:} Normal images(\textit{True label}) above and distorted images(\textit{False label}) below.}
  \label{fig:fig3}
\end{figure*}

From previous works, equivariance is defined as change in output according to corresponding change in input’s context. \finalEdit{Invariance is the opposite of equivariance where output does not change to change in input.} Figure~\ref{fig:2.4} simply shows concept of equivariance. In vision task, many researchers try to boost invariance to get better efficiency by using sub-sampling such as max-pooling. This way of boosting invariance makes model focus more on shape or texture, but gives limitation in understanding context such as objects' relations. To overcome this limitation, \aclStyle{Romero et al. (2020)} proposes “attentive group equivariant convolutions”~\cite{romero2020attentive} and shows that it has effect on equivariant training. In addition, \aclStyle{Sabour et al. (2017)} leads model to be more equivariant by applying "dynamic routing"~\cite{sabour2017dynamic}. From insights above, we believe that goal of VL tasks should lean towards equivariant model because goal of VL is to mimic human expressions, and we humans are equivariant. In this paper, We construct “\textit{Human Puzzle}” dataset which contains distorted context of human body and propose \textit{"Equivariance Score"} that can quantify the VL models’ equivariance.

\section{\changelines{Framework for Global Context and Equivariance}}
\changeliness{In Vision-Language Processing, understanding the global context of an image is an evident matter of emphasis for human-like understanding.} We suggest a simple approach to quantify how much a model is comprehending global context. Figure~\ref{fig:fig2} is an overview of our approach. Each image is forwarded to the VL model, where normal image is set to ~\textit{True}, while image with distorted global context \changelines{in human body} is set to ~\textit{False}.

The output of the model is then sent to a classifier instead of word embedding to distinguish its boolean value. Over this course, weights of the VL model are frozen, and only the classifier proceeds to train from the process. If a model is well capable of understanding global context, it would excel in classifying normal images \changeliness{from images with distorted human body, and this suggests the model is equivariant. If it can't classify well, then the model would be seen as invariant. In section~\ref{sec:6}, we suggest the quantifying metric of equivariance as ~\textit{Equivariance Score,} and this metric with ~\textit{Human Puzzle} dataset could reflect how model understands global context of human body. The above process is applicable to both VL models and vision models that take image as an input.}\\

\section{Dataset Construction}

\iffalse 모델이 equivariance한지 확인하는 가장 쉬운 방법은 context가 훼손된 이미지와 그렇지 않은 이미지를 모델에 입력하고 이를 비교하는 것이다. \fi
\changelines{Our suggestion to measure if a model is equivariant at global context is comparing the output of the normal image with that of image with distorted context.}
For \changelines{this process,} a complementary dataset is essential. To prepare normal images and images with distorted global contexts, we had to construct synthesized data. Basic concept originates from the fact that VL models do not fully understand human shape as shown in Figure~\ref{fig:fig1}. To acquire enough synthetic data, raw data with humans separated from the background was necessary. VideoMatte240K~\cite{lin2021real} dataset was selected as the source of raw data for human figures as it met such criteria. Total of 958 human figure images were collected by sampling the frame from VideoMatte240K dataset, which filmed humans in diverse actions without any background. In addition, 193 background images from PhotoMatte85 dataset were used as source of raw data.\\

\begin{table*}[t!]
\resizebox{\linewidth}{!}
{
\centering
\begin{tabular}{c c c c c}
\hline
\textbf{model} & \textbf{pre-trained dataset} & \textbf{entropy} & \textbf{acc.} & \textbf{\textit{equivariance score}}\\
\hline
Virtex~\cite{desai2021virtex} & COCO~\cite{chen2015microsoft} & 0.039 & 91.57\% & 0.736 \\
OSCAR+~\cite{li2020oscar} & COCO~\cite{chen2015microsoft} + $extra^{*}$ & 0.299 & 93.0\% & 0.740 \\
\textbf{X-VLM}~\cite{zeng2021multi} & COCO~\cite{chen2015microsoft} & 0.216 & \textbf{96.42\%} & \textbf{0.830} \\
\textbf{OFA}~\cite{wang2022unifying} & COCO~\cite{chen2015microsoft} & 0.0928 & \textbf{96.92\%} &  \textbf{0.914} \\ \hline
ViT~\cite{dosovitskiy2020image} & ImageNet-1K~\cite{imagenet_cvpr09} & 0.806 & 63.7\% & 0.473 \\
DViT~\cite{zhou2021deepvit} & ImageNet-1K~\cite{imagenet_cvpr09} &  0.678 & 73.72\% & 0.500 \\
SwinT~\cite{liu2021swin} & ImageNet-1K~\cite{imagenet_cvpr09} & 0.205 & 95.3\% & 0.821 \\
ResNet~\cite{he2016deep} & ImageNet-1K~\cite{imagenet_cvpr09}  & 0.096  & 95.5\% & 0.910 \\
SE-ResNet~\cite{hu2018squeeze} & ImageNet-1K~\cite{imagenet_cvpr09}  & 0.079 & 96.1\% & 0.920 \\\hline
\end{tabular}
}
\begin{flushright}
\footnotesize{
{$^*$ OSCAR+ follow the datasets settings of VinVL\cite{zhang2021vinvl}}}
\end{flushright}
\caption{Comparison of quantitative performance and \textit{Equivariance Score} for each model on \textit{Human Puzzle dataset}.}
\label{tab:1}
\end{table*}

\begin{table*}[t]
\resizebox{\linewidth}{!}
{
\centering
\begin{tabular}{c c c c c}
\hline
\textbf{model} & \textbf{pre-trained dataset} & \textbf{entropy} & \textbf{acc.} & \textbf{\textit{equivariance score}}\\
\hline
FCN~\cite{long2015fully} & Cityscapes~\cite{Cordts2016Cityscapes} & 0.538 & 89.1\% & 0.544 \\
% PSPNet~\cite{zhao2017pyramid} & Cityscapes~\cite{Cordts2016Cityscapes} & - & - & -\\
DeepLabV3~\cite{chen2017rethinking} & Cityscapes~\cite{Cordts2016Cityscapes} & 0.447 & 84.2\% & 0.633 \\
CCNet~\cite{huang2019ccnet} & Cityscapes~\cite{Cordts2016Cityscapes} & 0.382 & 84.3\% & 0.677 \\\hline
\end{tabular}}
\caption{\textit{Equivariance Score} performance comparison for CNN-based networks.}
\label{tab:2}
\end{table*}

\section{Human Puzzle Dataset}
Generating tens of millions of images by mix-and-matching 958 human images with 193 background images is \changeline{not difficult}. Yet, we cannot argue that the entirety of arbitrarily modified and synthesized image dataset is suitable for qualifying degree of interpretation on global context as \changeline{we suggested.} Hence, highly recognized vision models were used to select relevant and meaningful images from the pool of millions. The methods used to impair the global contexts of human images are simple. Methods of slicing, reflecting, relocating, etc. depending on arbitrary offsets were applied while maintaining the axis of the original image. Raw synthetic data reached 2.6M by combining background images with all possible numbers of cases of modification on original human images. To filter out the 2.6M images, ResNet, ViT~\cite{dosovitskiy2020image}, and SwinT~\cite{liu2021swin} were used. After training each model’s classifier while the individual weights were frozen, images that models could not accurately classify from a separate test set were chosen. Finalists for the dataset were selected from images where 2 out of 3 models gave incorrect classification. In the end, we manually filtered out the remaining images to create the \textit{Human Puzzle dataset}.\\

\section{Equivariance Score}\label{sec:6}
The aim of our research is not to classify distorted data, but to carry out regression analysis on the extent of global context interpretation. Hence, simply measuring accuracy would be inappropriate, while applying entropy would serve as an adequate metric to quantify a model's rate of confusion. We suggest an entropy-based metric like Equation ~\ref{eq:cc} to assess the model's global context comprehension level.

\begin{equation}
    H(p) = -plog_2p - (1-p)log_2(1-p),
    \label{eq:entropy}
\end{equation}
$p \in P$ refers to the probability value after applying softmax on ~\textit{Human Puzzle}-trained model’s logits. Applying Equation~\ref{eq:entropy}, we calculate the entropy of this probability $p$ to finally calculate \textit{Equivariance Score} written in Equation~\ref{eq:cc}.

\begin{equation}
    \textit{Equivariance Score} = \\ \frac{\sum_t(1-H(p_t))+\sum_f H(p_f)}{N_t + N_f},
    \label{eq:cc}
\end{equation}

In Equation~\ref{eq:cc}, $p_t$ and $p_f$ ($p_t, p_f \in P$) represent the probabilities of correct and incorrect prediction. Here, $N_t$ and $N_f$ refer to the number of $p_t$ and $p_f$ respectively. Through Equation~\ref{eq:cc}, the value of rightfully predicted results becomes substantial, while the value of wrongfully predicted results goes minimal. Theoretically, if all the predicted answers are correct, and entropy of the result is also zero, then a perfect score would be given.\\

\section{Experiment}
\subsection{Experiment Details}
For training of the \textit{Human Puzzle} dataset in this study, we used encoders of the pre-trained VL and vision model. In the case of VL, since most of the models use  transformers, the last hidden layer output was applied. In the case of CNN-based vision model, since ResNet backbone forms the majority, features of ResNet’s fourth stage were applied. In the case of DeepLabV3, and CCNet, we extracted from each model’s ASPP, and CC-attention module. We designed the extracted features from each of the models to be forwarded to independent individual fully-connected layers to carry out binary prediction. This was done to differentiate the features with distorted global context.\\

\subsection{Quantitative Result}
Table~\ref{tab:1} shows the result of the measured \textit{Equivariance Score} trained and evaluated \changeline{with \textit{Human Puzzle dataset}.} We proceeded with training and testing on the latest VL models and vision models. In the case of OSCAR+, \changeline{it was trained with tremendous amounts of data compared to other VL models and performed very well in most of VL tasks.} Nonetheless, OSCAR+ only extracts as many features as it can from images without processing relational information, leading us to predict relatively low \textit{Equivariance Score}, and this was found to be true based on our experiment results. In the case of X-VLM, it was aimed to apply global context to its training. By constructing inter-object relations as a separate dataset, \changeline{X-VLM} carried out step-by-step contrastive learning. Despite X-VLM \changeline{being trained} on a \changeline{much} smaller dataset compared to OSCAR+, it demonstrated a much better \textit{Equivariance Score}. Unified vocab suggested in OFA consists of not only the vocab’s text but also locational information or even image itself. This was implemented to convert transferred information to the model for better interpretation. \changeline{We expected that unified vocab will enhance global context interpretation,} and OFA in our experiment result indeed conveyed the highest \textit{Equivariance Score}.

We also proceeded on experimenting on vision models. In general, CNN-based models showed higher \textit{Equivariance Score} over transformer-based models. This can be seen as a result of structural characteristics of CNN, where features of the input layer are forwarded to the end layer.

Drastic difference between ViT and SwinT’s \textit{Equivariance Score} is implying this as well. ViT showed that it can apply images to transformer by slicing images into patches, and to overcome ViT’s shortcoming of inductive bias, SwinT applied shifted window structure ideated from CNN. Drawing from this, we expected SwinT’s \textit{Equivariance Score} to be higher than ViT’s, and this was indeed true, as SwinT’s \textit{Equivariance Score} was closer to CNN-based model, far exceeding that of ViT.

\changeline{Furthermore, we explored studies exploiting the global context in the field of vision to extensively validate the usefulness of our \textit{Human Puzzle dataset}. }\iffalse\textcolor{red}{Studies regarding the global context of an image are also under research in the field of vision. We looked to certify the usefulness of our \textit{Human Puzzle dataset} in various models.}\fi Table~\ref{tab:2} shows results of evaluating context score on semantic segmentation models. Amongst them, FCN did not put global context in consideration, and each model all used the ResNet backbone. FCN displayed the highest performance in classification on our \textit{Human Puzzle dataset}, but in contrast, scored the lowest on \textit{Equivariance Score}. Meanwhile, other models that utilized global context measured higher in \textit{Equivariance Score} compared to FCN. This result establishes that our \textit{Human Puzzle dataset} and our suggested \textit{Equivariance Score} are, indeed, valid.\\

\section{Conclusion}

In this paper, we propose the \textit{Equivariance Score} to measure \changeline{Vision-Language} model’s degree of  \changelines{equivariance and provide the \textit{Human Puzzle} dataset. These two can help future works with measuring how VL models are equivariant in global context.} Overall, models that attempted to extract information on global context measured relatively higher \textit{Equivariance Score} than models that did not, and our \textit{Equivariance Score} is a valid metric to indicate how much such attempts are meaningful. The proposed method in our paper is applicable not only to the fields of VL, but also in the fields of vision and can serve as a basis for research in many different domains.\\

\section{Limitation and Discussion}

\textit{Human Puzzle} dataset and \textit{Equivariance Score} metric suggested in this paper are not just restricted to VL, but in all fields that work with images. However, the dataset has its own limitations in that it only considers the global context of human  \changelines{body}. In consequence, there exists a need to construct appropriate dataset with distorted global context in both humans and diverse objects with backgrounds as well. Through this additional construction, metrics to quantify VL and vision models’ equivariance in global context can become more robust. Furthermore, \textit{Equivariance Score} suggested in this paper is a new measuring metric that is previously nonexistent, implying focus on global context for future research in VL and vision.

\bibliography{anthology,custom}

\begin{thebibliography}{26}
\expandafter\ifx\csname natexlab\endcsname\relax\def\natexlab#1{#1}\fi

\bibitem[{Antol et~al.(2015)Antol, Agrawal, Lu, Mitchell, Batra, Zitnick, and
  Parikh}]{antol2015vqa}
Stanislaw Antol, Aishwarya Agrawal, Jiasen Lu, Margaret Mitchell, Dhruv Batra,
  C~Lawrence Zitnick, and Devi Parikh. 2015.
\newblock Vqa: Visual question answering.
\newblock In \emph{Proceedings of the IEEE international conference on computer
  vision}, pages 2425--2433.

\bibitem[{Chen et~al.(2017)Chen, Papandreou, Schroff, and
  Adam}]{chen2017rethinking}
Liang-Chieh Chen, George Papandreou, Florian Schroff, and Hartwig Adam. 2017.
\newblock Rethinking atrous convolution for semantic image segmentation.
\newblock \emph{arXiv preprint arXiv:1706.05587}.

\bibitem[{Chen et~al.(2015)Chen, Fang, Lin, Vedantam, Gupta, Doll{\'a}r, and
  Zitnick}]{chen2015microsoft}
Xinlei Chen, Hao Fang, Tsung-Yi Lin, Ramakrishna Vedantam, Saurabh Gupta, Piotr
  Doll{\'a}r, and C~Lawrence Zitnick. 2015.
\newblock Microsoft coco captions: Data collection and evaluation server.
\newblock \emph{arXiv preprint arXiv:1504.00325}.

\bibitem[{Cordts et~al.(2016)Cordts, Omran, Ramos, Rehfeld, Enzweiler,
  Benenson, Franke, Roth, and Schiele}]{Cordts2016Cityscapes}
Marius Cordts, Mohamed Omran, Sebastian Ramos, Timo Rehfeld, Markus Enzweiler,
  Rodrigo Benenson, Uwe Franke, Stefan Roth, and Bernt Schiele. 2016.
\newblock The cityscapes dataset for semantic urban scene understanding.
\newblock In \emph{Proc. of the IEEE Conference on Computer Vision and Pattern
  Recognition (CVPR)}.

\bibitem[{Deng et~al.(2009)Deng, Dong, Socher, Li, Li, and
  Fei-Fei}]{imagenet_cvpr09}
J.~Deng, W.~Dong, R.~Socher, L.-J. Li, K.~Li, and L.~Fei-Fei. 2009.
\newblock {ImageNet: A Large-Scale Hierarchical Image Database}.
\newblock In \emph{CVPR09}.

\bibitem[{Desai and Johnson(2021)}]{desai2021virtex}
Karan Desai and Justin Johnson. 2021.
\newblock Virtex: Learning visual representations from textual annotations.
\newblock In \emph{Proceedings of the IEEE/CVF Conference on Computer Vision
  and Pattern Recognition}, pages 11162--11173.

\bibitem[{Dosovitskiy et~al.(2020)Dosovitskiy, Beyer, Kolesnikov, Weissenborn,
  Zhai, Unterthiner, Dehghani, Minderer, Heigold, Gelly
  et~al.}]{dosovitskiy2020image}
Alexey Dosovitskiy, Lucas Beyer, Alexander Kolesnikov, Dirk Weissenborn,
  Xiaohua Zhai, Thomas Unterthiner, Mostafa Dehghani, Matthias Minderer, Georg
  Heigold, Sylvain Gelly, et~al. 2020.
\newblock An image is worth 16x16 words: Transformers for image recognition at
  scale.
\newblock \emph{arXiv preprint arXiv:2010.11929}.

\bibitem[{He et~al.(2016)He, Zhang, Ren, and Sun}]{he2016deep}
Kaiming He, Xiangyu Zhang, Shaoqing Ren, and Jian Sun. 2016.
\newblock Deep residual learning for image recognition.
\newblock In \emph{Proceedings of the IEEE conference on computer vision and
  pattern recognition}, pages 770--778.

\bibitem[{Hu et~al.(2018)Hu, Shen, and Sun}]{hu2018squeeze}
Jie Hu, Li~Shen, and Gang Sun. 2018.
\newblock Squeeze-and-excitation networks.
\newblock In \emph{Proceedings of the IEEE conference on computer vision and
  pattern recognition}, pages 7132--7141.

\bibitem[{Huang et~al.(2019)Huang, Wang, Huang, Huang, Wei, and
  Liu}]{huang2019ccnet}
Zilong Huang, Xinggang Wang, Lichao Huang, Chang Huang, Yunchao Wei, and Wenyu
  Liu. 2019.
\newblock Ccnet: Criss-cross attention for semantic segmentation.
\newblock In \emph{Proceedings of the IEEE/CVF International Conference on
  Computer Vision}, pages 603--612.

\bibitem[{Jelinek et~al.(1977)Jelinek, Mercer, Bahl, and
  Baker}]{jelinek1977perplexity}
Fred Jelinek, Robert~L Mercer, Lalit~R Bahl, and James~K Baker. 1977.
\newblock Perplexity—a measure of the difficulty of speech recognition tasks.
\newblock \emph{The Journal of the Acoustical Society of America},
  62(S1):S63--S63.

\bibitem[{Li et~al.(2020)Li, Yin, Li, Zhang, Hu, Zhang, Wang, Hu, Dong, Wei
  et~al.}]{li2020oscar}
Xiujun Li, Xi~Yin, Chunyuan Li, Pengchuan Zhang, Xiaowei Hu, Lei Zhang, Lijuan
  Wang, Houdong Hu, Li~Dong, Furu Wei, et~al. 2020.
\newblock Oscar: Object-semantics aligned pre-training for vision-language
  tasks.
\newblock In \emph{European Conference on Computer Vision}, pages 121--137.
  Springer.

\bibitem[{Lin et~al.(2021)Lin, Ryabtsev, Sengupta, Curless, Seitz, and
  Kemelmacher-Shlizerman}]{lin2021real}
Shanchuan Lin, Andrey Ryabtsev, Soumyadip Sengupta, Brian~L Curless, Steven~M
  Seitz, and Ira Kemelmacher-Shlizerman. 2021.
\newblock Real-time high-resolution background matting.
\newblock In \emph{Proceedings of the IEEE/CVF Conference on Computer Vision
  and Pattern Recognition}, pages 8762--8771.

\bibitem[{Liu et~al.(2021)Liu, Lin, Cao, Hu, Wei, Zhang, Lin, and
  Guo}]{liu2021swin}
Ze~Liu, Yutong Lin, Yue Cao, Han Hu, Yixuan Wei, Zheng Zhang, Stephen Lin, and
  Baining Guo. 2021.
\newblock Swin transformer: Hierarchical vision transformer using shifted
  windows.
\newblock In \emph{Proceedings of the IEEE/CVF International Conference on
  Computer Vision}, pages 10012--10022.

\bibitem[{Long et~al.(2015)Long, Shelhamer, and Darrell}]{long2015fully}
Jonathan Long, Evan Shelhamer, and Trevor Darrell. 2015.
\newblock Fully convolutional networks for semantic segmentation.
\newblock In \emph{Proceedings of the IEEE conference on computer vision and
  pattern recognition}, pages 3431--3440.

\bibitem[{Pan et~al.(2020)Pan, Yao, Li, and Mei}]{pan2020x}
Yingwei Pan, Ting Yao, Yehao Li, and Tao Mei. 2020.
\newblock X-linear attention networks for image captioning.
\newblock In \emph{Proceedings of the IEEE/CVF Conference on Computer Vision
  and Pattern Recognition}, pages 10971--10980.

\bibitem[{Papineni et~al.(2002)Papineni, Roukos, Ward, and
  Zhu}]{papineni2002bleu}
Kishore Papineni, Salim Roukos, Todd Ward, and Wei-Jing Zhu. 2002.
\newblock Bleu: a method for automatic evaluation of machine translation.
\newblock In \emph{Proceedings of the 40th annual meeting of the Association
  for Computational Linguistics}, pages 311--318.

\bibitem[{Ren et~al.(2015)Ren, He, Girshick, and Sun}]{ren2015faster}
Shaoqing Ren, Kaiming He, Ross Girshick, and Jian Sun. 2015.
\newblock Faster r-cnn: Towards real-time object detection with region proposal
  networks.
\newblock \emph{Advances in neural information processing systems}, 28.

\bibitem[{Romero et~al.(2020)Romero, Bekkers, Tomczak, and
  Hoogendoorn}]{romero2020attentive}
David Romero, Erik Bekkers, Jakub Tomczak, and Mark Hoogendoorn. 2020.
\newblock Attentive group equivariant convolutional networks.
\newblock In \emph{International Conference on Machine Learning}, pages
  8188--8199. PMLR.

\bibitem[{Sabour et~al.(2017)Sabour, Frosst, and Hinton}]{sabour2017dynamic}
Sara Sabour, Nicholas Frosst, and Geoffrey~E Hinton. 2017.
\newblock Dynamic routing between capsules.
\newblock \emph{Advances in neural information processing systems}, 30.

\bibitem[{Vedantam et~al.(2015)Vedantam, Lawrence~Zitnick, and
  Parikh}]{vedantam2015cider}
Ramakrishna Vedantam, C~Lawrence~Zitnick, and Devi Parikh. 2015.
\newblock Cider: Consensus-based image description evaluation.
\newblock In \emph{Proceedings of the IEEE conference on computer vision and
  pattern recognition}, pages 4566--4575.

\bibitem[{Wang et~al.(2022)Wang, Yang, Men, Lin, Bai, Li, Ma, Zhou, Zhou, and
  Yang}]{wang2022unifying}
Peng Wang, An~Yang, Rui Men, Junyang Lin, Shuai Bai, Zhikang Li, Jianxin Ma,
  Chang Zhou, Jingren Zhou, and Hongxia Yang. 2022.
\newblock Unifying architectures, tasks, and modalities through a simple
  sequence-to-sequence learning framework.
\newblock \emph{arXiv preprint arXiv:2202.03052}.

\bibitem[{Wang et~al.(2021)Wang, Yu, Yu, Dai, Tsvetkov, and
  Cao}]{wang2021simvlm}
Zirui Wang, Jiahui Yu, Adams~Wei Yu, Zihang Dai, Yulia Tsvetkov, and Yuan Cao.
  2021.
\newblock Simvlm: Simple visual language model pretraining with weak
  supervision.
\newblock \emph{arXiv preprint arXiv:2108.10904}.

\bibitem[{Zeng et~al.(2021)Zeng, Zhang, and Li}]{zeng2021multi}
Yan Zeng, Xinsong Zhang, and Hang Li. 2021.
\newblock Multi-grained vision language pre-training: Aligning texts with
  visual concepts.
\newblock \emph{arXiv preprint arXiv:2111.08276}.

\bibitem[{Zhang et~al.(2021)Zhang, Li, Hu, Yang, Zhang, Wang, Choi, and
  Gao}]{zhang2021vinvl}
Pengchuan Zhang, Xiujun Li, Xiaowei Hu, Jianwei Yang, Lei Zhang, Lijuan Wang,
  Yejin Choi, and Jianfeng Gao. 2021.
\newblock Vinvl: Revisiting visual representations in vision-language models.
\newblock In \emph{Proceedings of the IEEE/CVF Conference on Computer Vision
  and Pattern Recognition}, pages 5579--5588.

\bibitem[{Zhou et~al.(2021)Zhou, Kang, Jin, Yang, Lian, Jiang, Hou, and
  Feng}]{zhou2021deepvit}
Daquan Zhou, Bingyi Kang, Xiaojie Jin, Linjie Yang, Xiaochen Lian, Zihang
  Jiang, Qibin Hou, and Jiashi Feng. 2021.
\newblock Deepvit: Towards deeper vision transformer.
\newblock \emph{arXiv preprint arXiv:2103.11886}.

\end{thebibliography}
\bibliographystyle{acl_natbib}

\end{document}